\documentclass{article}
\usepackage{ijcai18}

\usepackage{times}
\usepackage{xcolor}
\usepackage{soul}
\usepackage[utf8]{inputenc}
\usepackage[small]{caption}
\usepackage{caption}
\usepackage{graphicx} 
\usepackage{amsmath} 
\usepackage{ctable} 
\usepackage{epstopdf} 
\usepackage{subfigure} 

\title{Hybrid Approach of Relation Network and Localized Graph Convolutional Filtering for Breast Cancer Subtype Classification}

\author{
Sungmin Rhee$^1$, 
Seokjun Seo$^1$, 
Sun Kim$^{1,2,3}$ 
\\ 
$^1$ Department of Computer Science and Engineering, Seoul National University\\
$^2$ Bioinformatics Institute, Seoul National University, Seoul, Republic of Korea\\
$^3$ Interdisciplinary Program in Bioinformatics, Seoul National University\\
lars@snu.ac.kr,
dane2522@snu.ac.kr,
sunkim.bioinfo@snu.ac.kr
}
\begin{document}

\maketitle

\begin{abstract}
Network biology has been successfully used to help reveal complex mechanisms of disease, especially cancer. On the other hand, network biology requires in-depth knowledge to construct disease-specific networks, but our current knowledge is very limited even with the recent advances in human cancer biology. Deep learning has shown an ability to address the problem like this. However, it conventionally used grid-like structured data, thus application of deep learning technologies to the human disease subtypes is yet to be explored. To overcome the issue, we propose a hybrid model, which integrates two key components 1) graph convolution neural network (graph CNN) and 2) relation network (RN). Experimental results on synthetic data and breast cancer data demonstrate that our proposed method shows better performances than existing methods.
\end{abstract}

\section{Introduction}

Breast cancer is one of the most common cancers, especially leading type of cancer in women  \cite{cancer2012comprehensive}. Breast cancer has multiple risk factors for development and proliferation including genetic change, epigenetic change, and environmental factors. Also, breast cancer is a complex, multifactorial disease where interplays between these risk factors decide the phenotype of cancer such as progression, development, or metastasis. Thus, it is a challenging problem to determine how a cancer cell is developed and progressed.

Characterizing mechanisms of a complex disease as a whole is not possible. An effective approach is to define subtypes by dividing cancer into several categories according to various criteria such as phenotype, molecular portraits, and histopathology.
Among the several molecular properties based breast cancer subtypes, PAM50 \cite{parker2009supervised} has become a standardized model with the clinical utility to make diagnosis decisions in practice or building a treatment plan for a patient. Also, St. Gallen international expert consensus panel introduced a system for recommending adjuvant systemic therapy based on breast cancer subtype since 2011 \cite{goldhirsch2013personalizing}. However, despite the practical utility of breast cancer subtypes, they are still remained to be suboptimal since the complex mechanism underlying breast cancer cell is not fully investigated.

The main technical issue in elucidating biological mechanisms of breast cancer is that innate relational and cooperative characteristic of genes should be considered. In any specific biological context, multiple dysregulated genes derive phenotypic differences by mechanisms such as forming complexes, regulating each other, or affecting signal transduction. 
To address the technical difficulties related to complex associations, studies that utilize biological network are needed.
Early biological network studies have tried to discover distinct patterns based on edge information \cite{barabasi2004network,barabasi2011network}, and recent approaches rely on the common paradigm called \textit{network propagation}, which assumes that the information is propagated to nearby nodes through the edges \cite{cowen2017network}. 
However, network biology requires in-depth knowledge to construct disease-specific networks, but our current knowledge is very limited even with the recent advances in human cancer biology.

Deep learning has shown an ability to address the difficult situation like this. However, application of deep learning technologies to the classification of human disease subtypes is not straightforward since deep learning technologies conventionally use grid-like structured data and they are not mainly designed to handle graph data.
Recently, graph based deep learning techniques have emerged, which becomes an opportunity to leverage analyses in network biology.

In this paper, we try to advance network bioinformatics by incorporating a novel hybrid method of relation network (RN) and graph convolution neural network (graph CNN). Given the prior knowledge of putative associating genes represented in a graph structure, our proposed method captures localized patterns of associating genes with graph CNN, and then learn the relation between these patterns by RN.

The main contributions of this work are as follows:
\begin{itemize}
\item We propose a novel hybrid approach composed of graph CNN and RN. Our method is motivated by the fact that relations between entities are traditionally modelled with a graph structure. To the best of our knowledge, this work is the first of its kind.
\item We propose a model for biological networks. Since the dimension of the conventional biological network is large, we applied fast graph convolution filtering method that can scale up to the very large dimension. In addition, we modified the relation network to fit in the task.
\item We demonstrate the effectiveness of our approach by experiments using synthetic dataset and breast cancer subtype classification data. Our model is able to achieve good performance in terms of both classification evaluation and capturing biological characteristics such as survival hazard and subtype prognosis.
\end{itemize}

The article is organized as follows. 
In the next section, we review previous studies related to our work.
Then, our model is described in Section 3. In Section 4, we demonstrate the experimental result with synthetic and real dataset.

\section{Related Work}

\subsection{Deep Learnings on Graphs}
Recent survey papers \cite{niepert2016learning,bronstein2017geometric} present comprehensive surveys on graph deep learnings that recently emerge. In this section, we review a representative selection of the previous studies related to this work.

In the case of recurrent neural networks (RNN), there has been an attempt \cite{scarselli2009graph} to combine the graph structure with the neural network earlier than CNN. Graph neural network (GNN) is one of such study, which is an extension of recursive neural network and random walk. The representation of each node propagates through edges until it reaches a stable equilibrium. Then it is used as the features in classification and regression problems. This approach is further extended to a method named gated graph sequence neural network (GGS-NN), introducing gated recurrent unit and modifying to output sequence \cite{li2015gated}. 
Recently, \citeauthor{johnson2016learning} have proposed a method built upon GGS-NN by allowing graph-structured intermediate representations, as well as graph-structured outputs \cite{johnson2016learning}.

CNN has been successful on domains with underlying grid-like structured data such as computer vision, natural language processing, audio analysis, and DNA sequences. Recently, several works extended CNN to more general topologies like manifolds or graphs \cite{bruna2013spectral,henaff2015deep,niepert2016learning,defferrard2016convolutional}. \citeauthor{bruna2013spectral} have introduced a spectral formulation of graph CNN (SCNN), which draws on the properties of convolutions in the Fourier domain \cite{bruna2013spectral}. They extend the ideas to large-scale classification problems \cite{henaff2015deep}. Nevertheless, the method still does not scale up well due to the computational cost of $\Theta(n^2)$ for matrix multiplication.

The method proposed by \citeauthor{defferrard2016convolutional} leverage on the spectral graph CNN (SCNN), which is a key component of our approach. Computational efficiency is improved by using Chebyshev approximation technique. As a result, their method outperforms the existing SCNNs in terms of accuracy in their experiments.

Recently, Graph Attention Network (GAT) is proposed \cite{velivckovic2017graph}, which utilizes masked self-attention mechanism without convolution or RNN to deal with graph structured data. In the method, attention value is evaluated for each of neighboring nodes to produce feature vectors for graph nodes. GAT model has achieved the state-of-art results across four different experiments established on graph structured datasets.

\subsection{Relation Reasoning}
As Google DeepMind's team mentioned in their recent study \cite{santoro2017simple}, deductive reasoning methods innately reason about relations of entities from training data, which is also represented as relations \cite{quinlan1990learning}. However, these approaches lack the ability to deal with fuzzy and variational input data robustly \cite{harnad1990symbol}. Meanwhile in the statistical learning domain, DeepMind recently have proposed a method named relational network as a general solution to relational reasoning in neural networks \cite{santoro2017simple}.
The method focuses on relational reasoning with an easy-to-understand and generalizable network structure, which makes it easy to be modified or combined with other methods. In addition, despite its simplicity in structure, it has demonstrated super-human performance in visual question answering problem. In this paper, we modify the relation network and combine it with graph CNN for the first time. It shows that the relational reasoning helps to improve the performance of the proposed breast cancer subtype classification task.

\begin{figure*}[!t]
\centering
\includegraphics[width=6in]{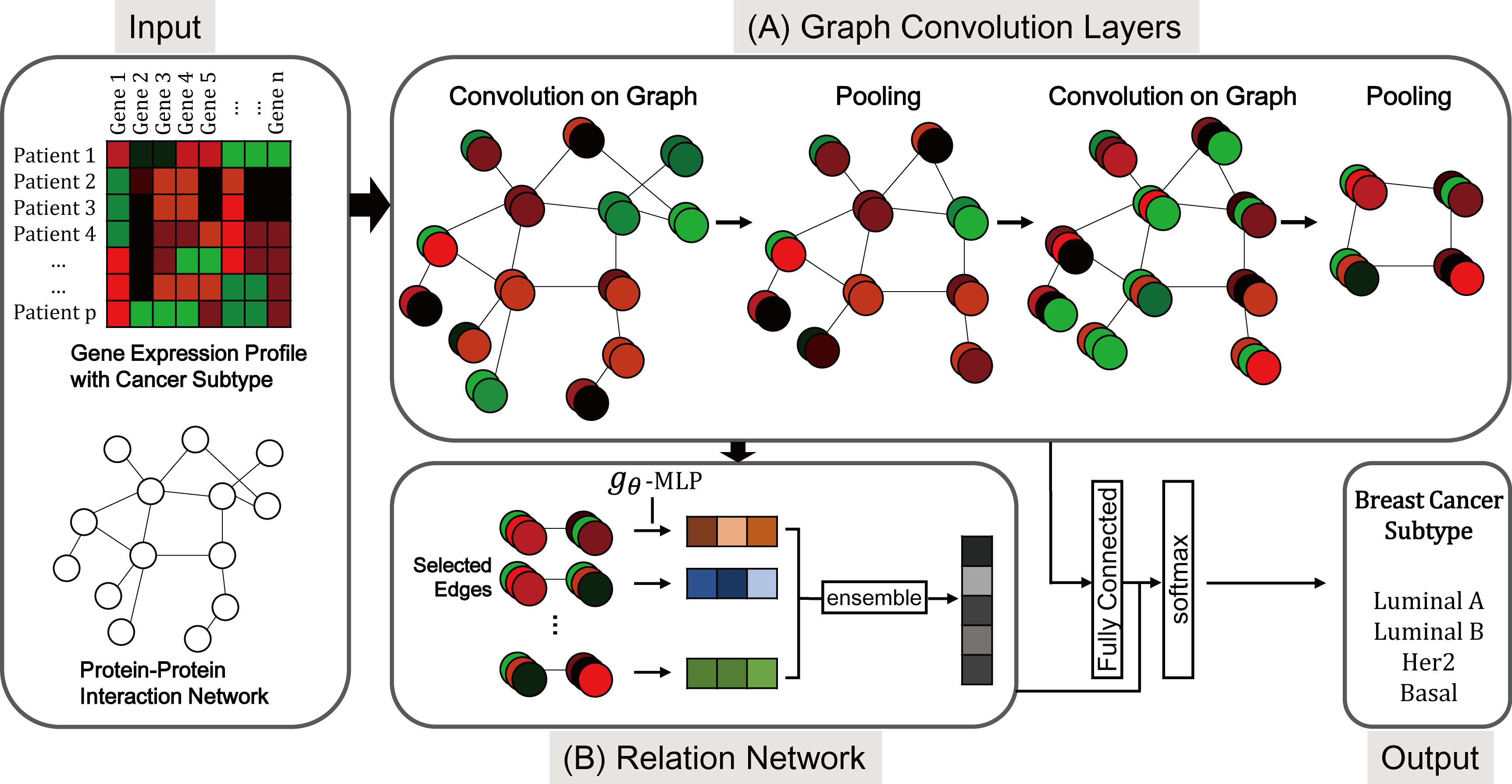}
\caption{Overview of the proposed method}
\label{fig_workflow}
\end{figure*}

\section{Methods}
In this section, we describe the proposed method. Figure \ref{fig_workflow} illustrates the overall workflow of the proposed method.
The first step (Figure \ref{fig_workflow} A) of the method is the graph convolution step to represent and capture localized patterns of the graph nodes (genes). The second step (Figure \ref{fig_workflow} B) is the relational reasoning step. In this step, the model learn the complex association between graph node groups (gene sets) from the learned localized patterns of graph nodes (genes) in the previous step. The next step is to merge the representation of graph convolution layer and relation reasoning layer.

In this paper, we will denote data elements of each sample $p$ as $x_p \in R^n$, and weighted graph topology as $G = (V,E,A)$, where $V$ and $E$ represent the sets of vertices and edges, respectively. Also, we will use $A$ to denote the weighted adjacency matrix and $N$ to denote the number of vertices, i.e. $|V|$.

\subsection{Localized Pattern Representation by Graph Convolution Neural Network}
For capturing localized patterns of data (gene expression profile), we first mapped input data in the graph structure and used graph CNN technique to find localized patterns of the graph signal.
Let $x_p \in R^{n}$ be the graph signal (or gene expression) in the sample $p$.
Then graph Laplacian matrix $L$ is used to find spectral localized patterns of $x_p$ under the graph structure $G$. Laplacian matrix $L$ of graph $G$ is defined as $L = D - A$ where $D$ is a weighted degree matrix, and $A$ is a weighted adjacency matrix of graph $G$. 

Then the graph convolution of signal $x$ is defined with graph Laplacian matrix. Let's assume that $L = U\Lambda U^T$ is an eigenvalue decomposition of graph Laplacian $L$, where $U = [u_1, ... , u_n]$ is a matrix composed of eigenvectors $\{u_l\}_{l=1}^{n}$ and $\Lambda$ is a diagonal matrix $diag([\lambda_1,...,\lambda_n])$ composed of eigenvalues $\{\lambda_l\}_{l=1}^{n}$. We can say that  $\{u_l\}_{l=1}^{n}$ is a complete set of orthonormal eigenvectors as $L$ is a symmetric positive semidefinite matrix \cite{defferrard2016convolutional}. Then the graph Fourier transform is defined as $\hat{x} = U^Tx$ and inverse graph Fourier transform is defined as $x = U\hat{x}$ \cite{shuman2013emerging}.

Unlike in the classical signal processing domain, it is not straightforward to define the convolution of two signals in spectral graph domain. Thus, convolution theorem in equation \ref{eq_convolution_theorem} is borrowed from classical signal processing domain to define graph convolution as follows,

\begin{equation}
\label{eq_convolution_theorem}
(x*y)(t) =  F^{-1}(F(f)F(g))
\end{equation}

\noindent where $F$ and $F^{-1}$ denotes Fourier and inverse Fourier transform for each, and $x$, $y$ denotes two input signals. From the definition of Fourier transform in graph spectral domain, we can induce graph convolution by combining convolution theorem and graph Fourier transform as following equation

\begin{align}
\label{eq_graph_convolution}
x*_{G}y & = U( (U^Tx) \odot (U^Ty)) \nonumber \\
& = U ( (U^Ty) \odot (U^Tx)) \nonumber \\
& = U y(\Lambda) U^Tx
\end{align}

\noindent where $\odot$ is the element-wise Hadamard product and $y(\Lambda) \in R^n$ is a diagonal matrix $diag([\hat{y}(\lambda_1),...,\hat{y}(\lambda_n)])$. Since the matrix $U$ is determined by the topology of input graph and invariant, only the matrix $y(\Lambda)$ determines various forms of convolution filters, i.e, elements in $y(\Lambda)$ matrix only are learnable parameters in graph convolution.
Among several possible graph convolution filters, our method used polynomial parametrized filter $y_{\theta}(\Lambda) = \sum_{k=0}^{K-1}\theta_{k}\Lambda^{k}$ that can express localized signal patterns in $K$-hop neighboring nodes.
However, evaluating polynomial parametrized filter requires very expensive computational complexity $O(n^2)$. To deal with this circumstance, previous study \cite{hammond2011wavelets} proposed an approximated polynomial named Chebyshev expansion. The Chebyshev polynomial $T_k(x)$ of order $k$ is recursively defined as $T_k(x) = 2xT_{k-1}(x) - T_{k-2}(x)$ with $T_0 = 1$ and $T_1 = x$. Then the filter can be approximated as $y_{\theta'}(\Lambda) = \sum_{k=0}^{K-1} \theta_k'T_k(\tilde{\Lambda})$ with $\tilde{\Lambda} = 2\Lambda/\lambda_{max} - I_n$.

Going back to the graph convolution in equation \ref{eq_graph_convolution}, we can now define the final graph convolution filter as
\begin{equation}
\label{eq_cheybyshev}
x*_Gy_{\theta'} = \sum_{k=0}^{K-1}\theta_k'T_k(\tilde{L})x
\end{equation}
\noindent where $\tilde{L} = 2L/\lambda_{max} - I_n$ is a rescaled graph Laplacian. The equation \ref{eq_cheybyshev} can be easily induced from the observation $(U\Lambda U^T)^k = U\Lambda^kU^T$. Note that the entire filtering operation only requires $O(K|E|F_{in}F_{out}P + K|V|)$ and is fully differentiable, where $F_{in}$ and $F_{out}$ are the number of input and output convolution filters. Thus, entire filtering operation can be learned by backpropagation algorithm \cite{defferrard2016convolutional}.

Convolutioned graph signal is further pooled with neighboring nodes identified by Graclus algorithm \cite{dhillon2007weighted} as proposed in the previous study \cite{defferrard2016convolutional}. There are several pooling strategies in neural network such as max pooling and average pooling. Empirically, the average pooling performed best in our experiments. Therefore, we used average pooling in the proposed method.

\subsection{Learning Relation Between Graph Entities Using Relation Network}
To reason about association between graph nodes, we use relation network (RN), originally defined as

\begin{equation}
\label{eq_vanilla_RN}
RN(O) = f_{\phi}(\sum_{i,j}g_{\theta}(o_i,o_j))
\end{equation}

\noindent in the previous study \cite{santoro2017simple} where the input is a set of objects $O = \{o_1, o_2,..., o_n\}, o_i \in R^m$.
In the original study by \citeauthor{santoro2017simple}, multi-layer perceptron (MLP) is used for function of $f$ and $g$, and the parameters $\theta$ and $\phi$ are synaptic weights of perceptrons.
Unlike the task in this paper, there also exists a query for each of the input samples, and each query is embedded to the vector $q$ by an LSTM. Then, \citeauthor{santoro2017simple} re-define the RN architecture as $RN(O) = f_{\phi}(\sum_{i,j}g_{\theta}(o_i,o_j,q))$ so that it can process the query in the neural network.
This query embedding $q$ can work similarly as an attention mechanism. In other words, query embedding has an ability to select object pairs that are important for the classification task. Also, coordinate values of objects are used to define object pairs in the work of \citeauthor{santoro2017simple} as it takes an image input, which has innate coordinate information. Therefore, even if all pairs of objects are considered in the original RN, it is able to achieve a good performance.

\begin{figure}
\centering     
\subfigure[Original RN]{\label{fig:a}\includegraphics[width=60mm]{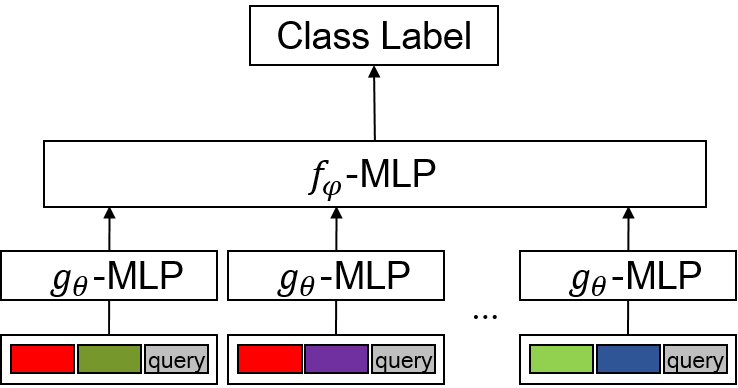}}
\subfigure[RN in the proposed method]{\label{fig:b}\includegraphics[width=60mm]{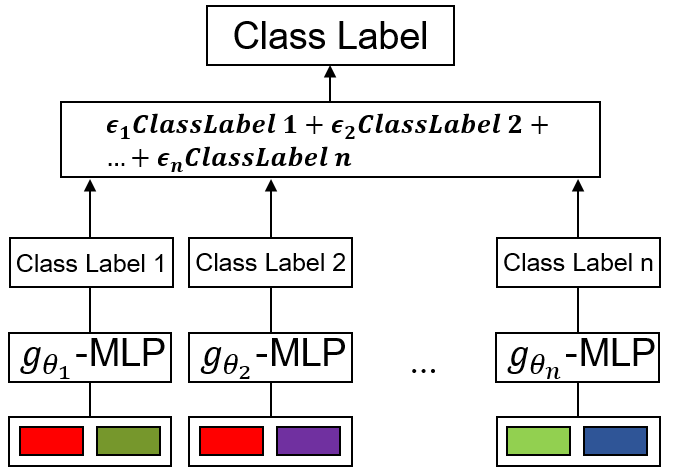}}
\caption{Architectures of relation networks}
\label{fig_RN_diff}
\end{figure}

However, no query information is available in our task and the number of considering objects is larger than the original work. This leads to two technical problems for relation reasoning, 1) object pairs that are not relevant to solve the problem can interfere with learning, and 2) considering all pairs is not feasible as the number of the objects is too large.

To deal with this problem, we have modified the relation network to fit in our task. At first, we sort the edges in the descending order of edge weights. Then top $\kappa$ number of edges are selected as input object pairs to effectively reduce the model size and exclude object pairs that are not highly relevant with the classification problem. Relations in each of the selected pairs are then inferred with $g$ function likewise in original relation network. Unlike the original relation network however, we used different kinds of functions for $g$ and $f$. We use MLPs with separated parameters for each of object pairs other than one MLP with shared parameters as $g$ function. In this way, g functions can learn different relations for each of object pairs. Also, linear ensemble summation is used for $f$ function. The different architectures of original RN and RN in the proposed method are illustrated in Figure \ref{fig_RN_diff}.

The final object relation network is inferred as 

\begin{equation}
\label{eq_brca_RN}
RN(O) = \sum_{i,j} \epsilon_{ij}g_{\theta_{ij}}(o_i, o_j)
\end{equation}

\noindent where $\epsilon_{i,j}$ is a learnable parameter that can be interpreted as an attention of each object pairs for the task, $o_i \in R^m$ is object embedding for each of graph nodes, and $m$ is the number of convolution filter in the last graph convolution layer.

\subsection{Merging Graph Convolution Layer and Relation Network}
The final output of the proposed model architecture is defined as 
\begin{equation}
\label{eq_network_output}
\hat{y} = softmax(h(x_i) + \sum_{i,j} \epsilon_{ij}g_{\theta_{ij}}(o_i, o_j))
\end{equation}

\noindent to combine outputs from graph convolution layer and relation network, where $h(x_i)$ is defined as a composition of functions; graph convolution, pooling, and fully connected layer.

\begin{figure}[!t]
\centering
\includegraphics[width=3.2in]{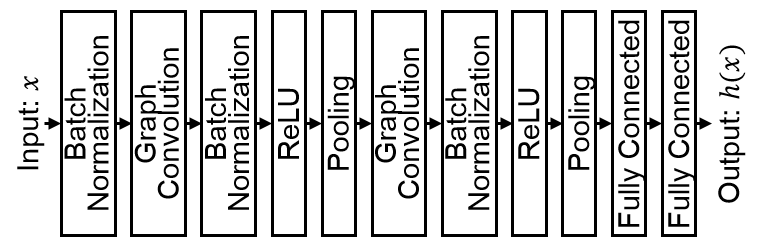}
\caption{Illustration of $h$ function}
\label{fig_h}
\end{figure}

To be more specific about $h$, we summarize the procedure of $h$ as follows. First, the input signal $x_i$ is normalized by a batch normalization method \cite{ioffe2015batch} to make learning process stable since our dataset has large absolute value and variance. 
Then, the normalized input signal is filtered by a graph convolution layer as defined in equation \ref{eq_cheybyshev}. Next, the convoluted signal is normalized through a batch normalization method so that the learning process can be accelerated and have regularization effect.
Then, ReLU activation function and average pooling are applied. 
We named the procedure from graph convolution to average pooling as \textit{graph convolution layer}.
After two graph convolution layer, a final feature map is used as an input of fully connected layer. Function $h$ is illustrated in Figure \ref{fig_h}.

Output at the last graph convolution layer is also used as a input of the relation network. 
Then cross-entropy between $\hat{y}$ in equation \ref{eq_network_output} and classification label is minimized by Adam \cite{kingma2014adam} algorithm.

Hyperparameters for learning procedure is determined as follows. Two graph convolution layer were used. Each layer has 32 convolution filters. $K$ of the first layer is 10 and second layer is 2. Pooling size is 2 for both of the layers. Two fully connected layers are used with 1024, 512 hidden nodes for each.
In the relation network, top 200 and 1000 edges were selected for each of synthetic and real dataset experiment. MLPs for $g$ function have one layer with 128 hidden nodes in synthetic dataset, and 2 layers with 128 hidden nodes for real dataset.

\section{Results and Discussion}

For comparison of performance, we used several methods in experiments of synthetic and real dataset.
GCNN denotes graph convolution neural network that has identical hyperparameters with the proposed method. GCNN+RN denotes simple integration of GCNN and vanilla relation network. As we mentioned in methods section, we modify the RN to fit in our task, and GCNN+RN uses original relation network to confirm the effectiveness of RN modification. GAT denotes graph attention network, which is the state-of-the-art neural network that deals with graph structured data. We also compared the method with several baseline methods; RF (Random Forest), kNN (k Nearest Neighbor), SVM (Support Vector Machine), MNB (Multinomial Naive Bayesian), and GNB (Gaussian Naive Bayesian). GNB is used for synthetic dataset since the data is sampled using Multinomial Gaussian Distribution, and MNB is used for real dataset since gene expression data is known to follow multinomial distribution. In real dataset experiment, we also used recent breast cancer classification method based on ensemble SVM \cite{huang2017svm}. 
The classification performances were measured by a monte-carlo cross validation experiment. We repeatedly sampled 10\% of samples as a validation set and used remaining 90\% of samples as a training set. For each of data splits, the proposed
model was fit to training set and accuracy was assessed using validation set. The accuracies were averaged over data splits.

\subsection{Synthetic Experiment}
\begin{figure}[!t]
\centering
\includegraphics[width=3.2in]{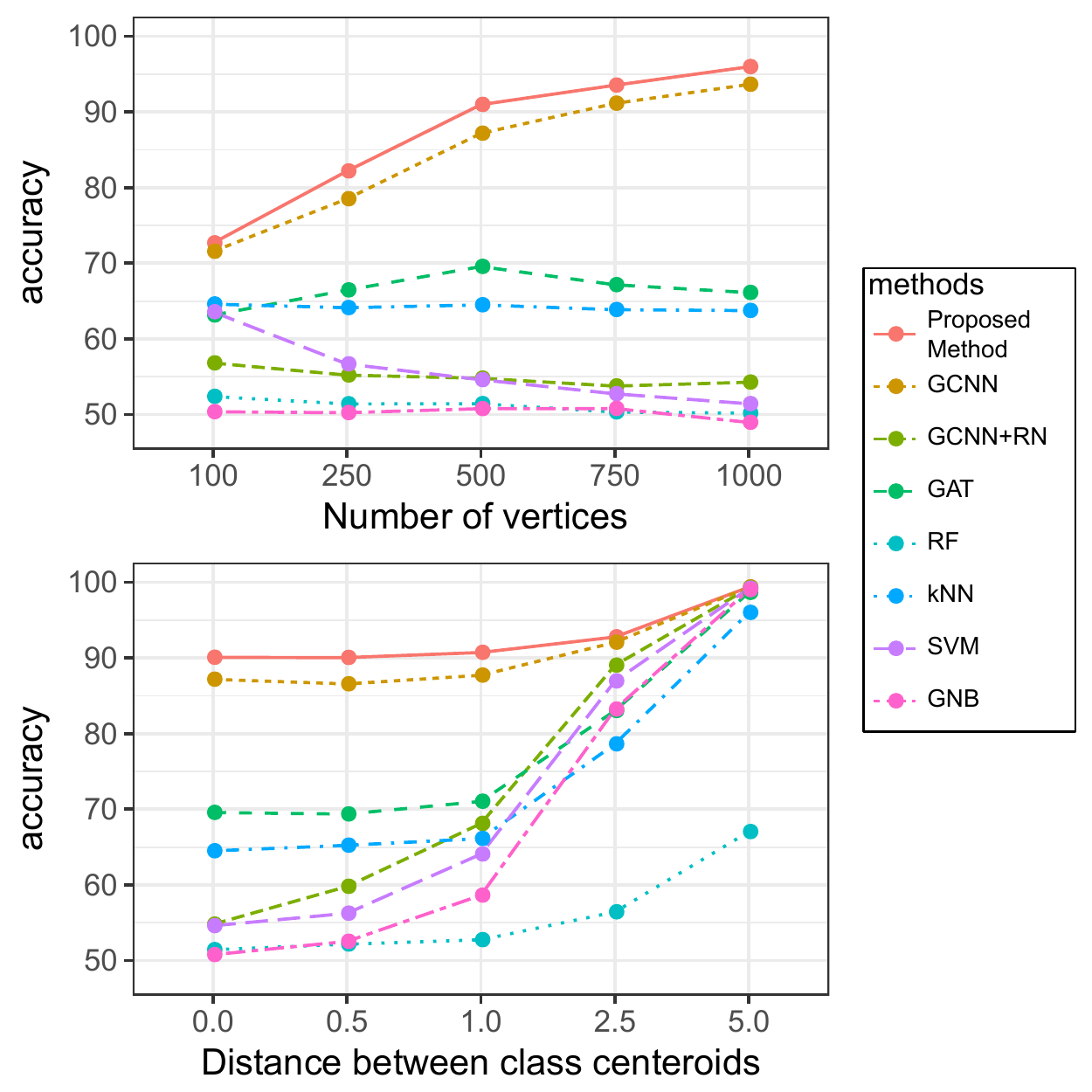}
\caption{Synthetic data performance comparison. Top plot shows performances across the number of vertices, and bottom plot shows performances across distance between class centroids.}
\label{fig_synthetic}
\end{figure}

The first experiment use randomly generated data with different underlying graph structure across classes. The experiment aims to test the ability of model when there are dependencies among attributes (graph vertices) and overlaps among different classes. We used MND (Multivariate Normal Distribution) to generate two class dataset, and each of classes has different covariance matrix to represent different underlying graph structure. We first generated template covariance matrix $\Sigma$. $\Sigma$ is sampled to have average degree 10, and each element in $\Sigma$ is sampled with normal distribution, i.e., $\sigma_{ij} \sim \mathcal{N}(0.1, 0.1)$ where $\sigma_{ij}$ is an element in $i$'th row and $j$'th column of $\Sigma$. Then the covariance matrices of two classes, $\Sigma^1$ and $\Sigma^2$, are generated by randomly deleting or reversing the sign of elements in template covariance matrix. In other words, $\sigma_{ij}^k = \mathcal{U}\{-1,1\} \cdot \sigma_{ij}, k \in \{1,2\}$, where $\mathcal{U}$ is the discrete uniform distribution and $k$ is an index for two classes. For $\mu$ in MND of each classes, we both test identical ones and two $\mu^1$, $\mu^2$ with small euclidean distances.

Figure \ref{fig_synthetic} shows that our method performs best in all of the experiments except one case with the largest centroid distance. Even in the experiment with the largest centroid distance, performance difference is marginal (99.35\% by the proposed method, 99.36\% by GCNN). 
Note that the synthesized dataset is difficult to be classified by approaches like finding hyperplane, since there are large overlap among data points in two classes. Thus, inferring relation between attributes is essential to classify the designed synthetic data. The experiment shows the effectiveness of our method, when dataset that have dependencies among attributes is need to be analysed.

\subsection{Real Dataset}
We applied the proposed method on dataset of human breast cancer patient samples. RNA-seq based expression profiles of genes are extracted from TCGA breast cancer level 3 data \cite{prat2012comprehensive}.
There are 57,292 genes in the original expression profile, and we excluded genes that were not expressed and further selected 4,303 genes in the cancer hallmark gene sets  \cite{liberzon2015molecular} to utilize only genes that are relevant with tumor.

For the classification label of the patient, PAM50 molecular subtype is used. PAM50 is the most commonly used breast cancer subtype scheme. The subtype includes Luminal A, Luminal B, Basal-like, and HER2.
Luminal subtype cancer cells are mostly grown from inner (luminal) cells of mammary ducts and known to have better prognoses than other subtypes. Compared to Luminal A however, Luminal B subtype tumors tend to have poorer prognosis factors like
higher tumor grade, larger tumor size, and lymph node involvement. Basal-like cancer cells are mostly grown from outer (basal) cells of mammary ducts and known to have worst prognoses and survival rates. HER2 subtype had its name since most HER2 subtype tumors are HER2-positive. HER2 subtype tumors tend to have poorer prognoses than luminal subtype tumors. In our study, 338 Luminal A, 265 Luminal B, 149 HER2, and 231 Basal-like patient samples were used for the experiment.

For the topology of the graph, we used STRING protein-protein interaction network \cite{szklarczyk2014string}. STRING is a curated database of putatively associating genes from multiple pieces of evidence like biological experiments, text-mined literature information, computational prediction, etc.

\subsection{Comparison of Classification Performance}

\begin{table}[!t]
\renewcommand{\arraystretch}{1.3}
\centering
\setlength\tabcolsep{3pt}
\begin{tabular}{|c|c|c|c|c|}
\hline
Methods&Peak&Final&F1&F1\\
&accuracy&accuracy&(support)&(Macro)\\
\specialrule{.1em}{.05em}{.05em}
\specialrule{.1em}{.05em}{.05em}
\textbf{Proposed}&\textbf{86.29\%}&\textbf{83.19\%}&\textbf{83.41\%}&\textbf{82.26\%}\\
\textbf{Method}&&&&\\
GCNN+RN&67.65\%&62.76\%&62.27\%&59.88\%\\
GCNN&85.27\%&82.39\%&82.52\%&81.26\%\\
GAT&85.64\%&81.37\%&81.12\%&80.15\%\\
SVM&-&77.53\%&77.53\%&76.11\%\\
MNB&-&75.45\%&75.87\%&77.53\%\\
RF&-&78.54\%&78.14\%&75.87\%\\
kNN&-&62.53\%&60.22\%&57.72\%\\
Huang \it{et al.}&-&71.68\%&70.70\%&67.93\%\\
\hline
\end{tabular}
\caption{Performance comparison of the methods on breast cancer subtype classification}
\label{accuracy_table}
\end{table}

Table \ref{accuracy_table} lists accuracies of the proposed model and comparing methods. Peak accuracies during learning processes were listed for top 4 methods in the table, and final accuracies after learning are listed for all of the comparing methods. Also F1 score with two average scheme, weighted by support (the number of true instances for each label) and Macro averaged, are listed. We can see that the proposed method performs best. Also, the simple integration of graph CNN and vanilla RN (GCNN+RN) shows the second worst performance. We use identical hyperparameters with the proposed method for GCNN+RN. We believe that GCNN+RN performs poor since, as we described in the method section, the original RN gets query encodings and coordinate values as inputs, that can work as a clue for relevant object selection. However, as there is no coordinate value or query in our task, the changes in our hybrid approach is efficient to make an increase in performance. 

\subsection{Consistency of tSNE Visualization and PAM50 Subtype Prognosis}

\begin{figure}[!t]
\centering
\includegraphics[width=3.35in]{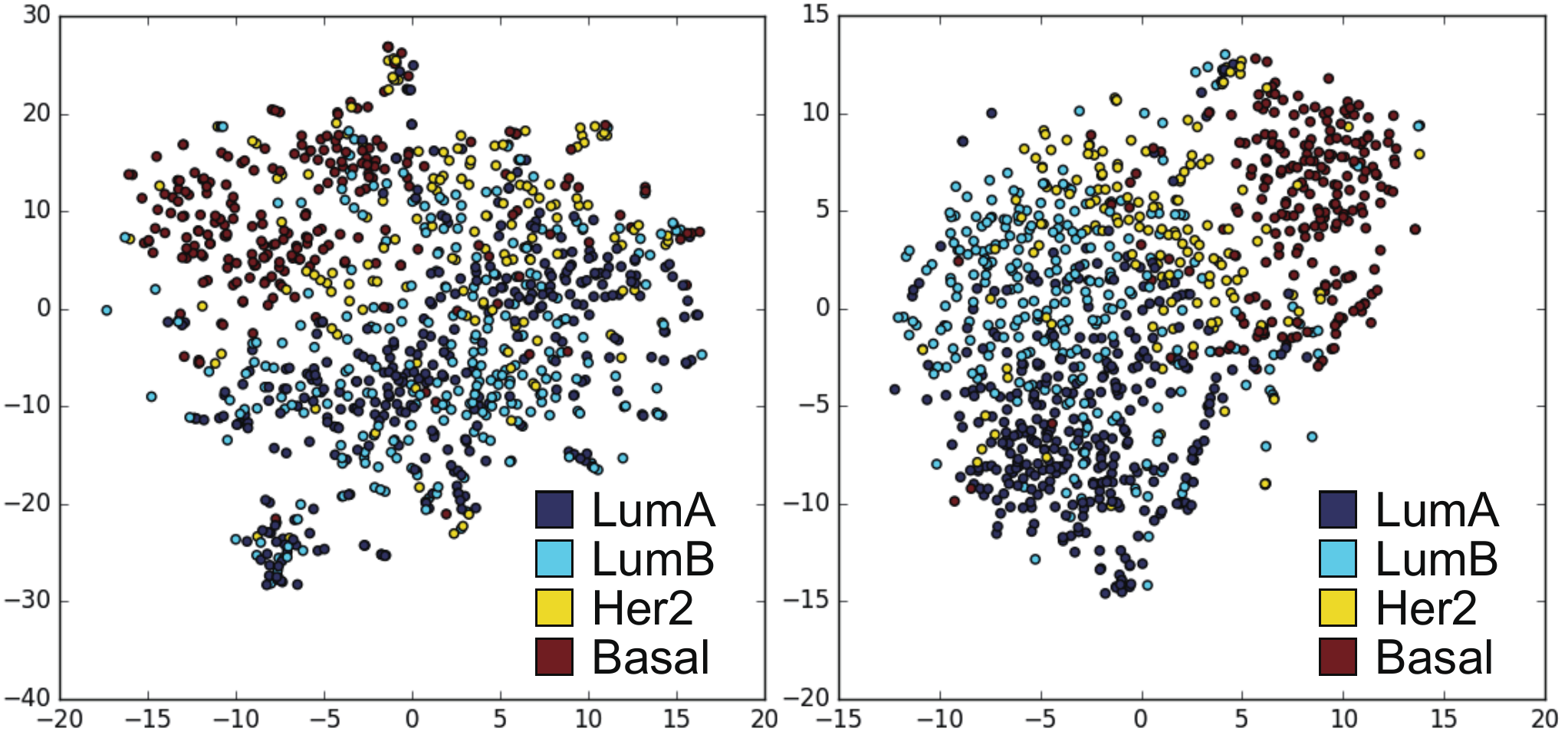}
\caption{tSNE visualization of graph convolution feature map}
\label{fig_tSNE}
\end{figure}

To qualitatively study whether the learned representation can express the biological characteristic of the patients, tSNE plot \cite{maaten2008visualizing} of the last convolution feature map is drawn (Right of Figure \ref{fig_tSNE}). Only the representation vectors of the objects, which are inputs of relation network, are used to plot. Output of RN is not used since it indicates class labels itself. We can see distinctive patterns between four subtype patients in the plot. However, the distinction between subtypes is not clear than typical examples e.g., tSNE plot of MNIST handwritten digits. We believe that this shows the complexity of the problem we are solving in the task. As we described earlier, the problem has higher input dimension and association between each feature should be considered.

More interestingly, we can see that the order of subtypes in the tSNE plot is identical to the order of prognosis of breast cancer subtypes. It is a well-known fact in the breast cancer clinical domain that Basal-like subtype has the worst prognosis, followed by HER2, Luminal B, and Luminal A. Especially, Basal-like subtype is known to have distinctive molecular characteristics from other subtypes \cite{bertucci2012basal}, which is also represented in Figure \ref{fig_tSNE}. All of these patterns is not significant in the tSNE plot with raw gene expression (left of Figure \ref{fig_tSNE}). Thus, we can say that the proposed method successfully learn the latent molecular properties in the expression profile of the patient samples.

\subsection{Survival Analysis}

\begin{figure}[!t]
\centering
\includegraphics[width=3.2in]{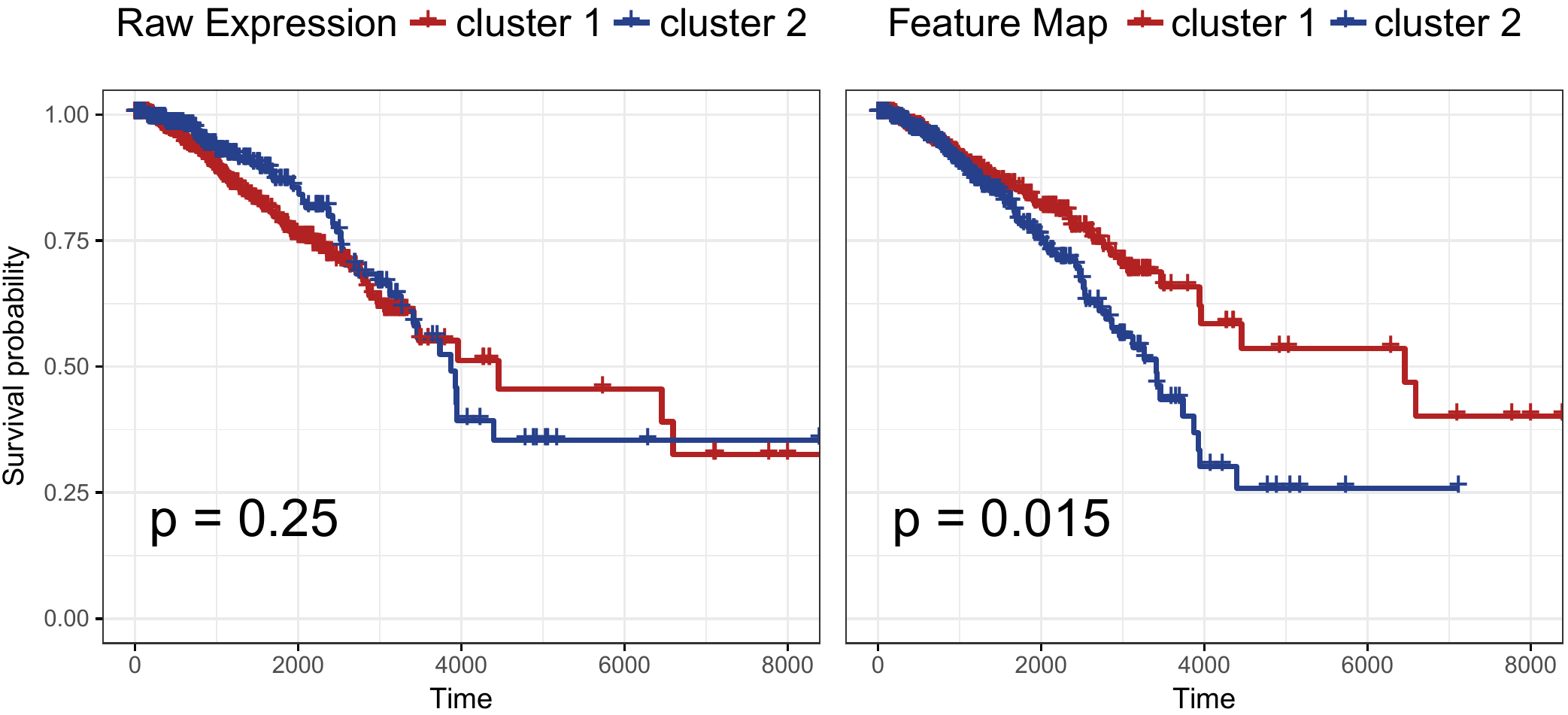}
\caption{Kaplan meier survival plot of patients}
\label{fig_KM}
\end{figure}

To further evaluate the ability of the model to comprehend characteristics of molecular subtypes, we performed survival analysis. 
We clustered the patients into two groups based on raw gene expression values and feature map data at the last graph convolution layer with dimensions reduced.
Agglomerative hierarchical clustering with Ward's criterion \cite{ward1963hierarchical} is used for clustering and tSNE is used for dimension reduction. Then Kaplan-meier plots (KM plot) \cite{kaplan1958nonparametric} drawn for each of two clustering results are seen in Figure \ref{fig_KM}. 
KM plot is standard analysis using non-parametric statistics to measure hazard ratios of different patient groups. In medical science, KM plot is often used to analyze the effectiveness of treatment by comparing KM plot of treated and non-treated patient groups.

The plot generated by feature map values (Right of Figure \ref{fig_KM}) shows that the patient samples are successfully divided into two subgroups that have distinct survival patterns with a p-value smaller than 0.05, while the plot with raw expression value (Left of Figure \ref{fig_KM}) failed. This is an interesting result as it shows that the model can simultaneously learn the phenotypic information such as prognosis of the patient while performing the classification task, which is not directly related with the information.

\section{Conclusion}
In this study, we show that hybrid approach of relation network and graph convolution neural network can learn the complex molecular mechanisms underlying breast cancer cells.
The proposed method is designed to perceive cooperative patterns of genes and their associations.
We observed that the method is successful to capture molecular characteristics of breast cancer in both quantitative and qualitative evaluation.
We anticipate that our approach can extend the territory of both network bioinformatics and deep learnings.
One important future work of the method is to extend the model to manage multiple heterogeneous data sources like sRNA sequencing, DNA methylation, as well as gene expression data. To do this, we plan to extend the model by incorporating other techniques such as multi-view learning and/or transfer learning.

\section*{Acknowledgments}

This research is supported by Next-Generation Information Computing Development Program through the National Research Foundation of Korea(NRF) funded by the Ministry of Science, ICT(No.NRF-2017M3C4A7065887),
the Collaborative Genome Program for Fostering New Post-Genome Industry of the National Research Foundation (NRF) funded by the Ministry of Science and ICT (MSIT) (No.NRF-2014M3C9A3063541),
and a grant of the Korea Health Technology R\&D Project through the Korea Health Industry Development Institute (KHIDI), funded by the Ministry of Health \& Welfare, Republic of Korea (grant number : HI15C3224).

\bibliographystyle{named}
\bibliography{myRef}

\end{document}